\documentclass[conference]{IEEEtran}
\usepackage[T1]{fontenc}
\usepackage{newtxtext,newtxmath}
\usepackage{booktabs}
\usepackage{graphicx}
\usepackage{cite}
\usepackage{amsmath}
\usepackage{url}
\usepackage[hidelinks]{hyperref}
\renewenvironment{abstract}{\small\bfseries\noindent\textit{Abstract:}\ }{\par}
\renewenvironment{IEEEkeywords}{\small\bfseries\noindent\textit{Index Terms:}\ }{\par\vspace{4pt}}

\begin{document}

\title{The Indic Token Tax in 2026: Measuring Tokenizer Disparity and Its Cost for the Languages of India}

\author{\IEEEauthorblockN{Johin Johny Arimbur}
\IEEEauthorblockA{Havon AI\\
Mumbai, India\\
johinjohny144@gmail.com}
\and
\IEEEauthorblockN{Ankur Ashit Ghosh}
\IEEEauthorblockA{Havon AI\\
Mumbai, India\\
ankurghosh2910@gmail.com}
\and
\IEEEauthorblockN{Shervin Sam}
\IEEEauthorblockA{Havon AI\\
Mumbai, India\\
samshervin37@gmail.com}}

\maketitle

\begin{abstract}
Commercial language model APIs bill per token, so a tokenizer that needs more tokens for the same content in one language makes that language more expensive, slower, and able to fit less text in a fixed context window. We measure this disparity for nine public tokenizers, including those of GPT-4, GPT-4o, Qwen2.5, DeepSeek-V3, Mistral-Nemo and four Sarvam models, on all 25 languages of India available in FLORES-200 (2,009 professionally translated sentences aligned with English) and on 13,732 human-translated Hinglish-English social media pairs from PHINC. On GPT-4o's tokenizer the same content needs between 1.57 times (Hindi) and 5.01 times (Odia) the tokens of English, and 13.7 times for Santali in Ol Chiki script; GPT-4's tokenizer needs up to 12.8 times. Sarvam-30B/105B is the only third-party tokenizer that stays below 2.6 times on every language (median 1.53). In an illustrative scenario using official list prices accessed on 5 November 2026 and stated assumptions, a support bot handling 100,000 chats a month on GPT-4o would cost about INR 56,640 in English and INR 102,829 in Marathi; these are estimates, not measured production costs. Against an openly licensed Hindi and Marathi morphological segmentation set, no tokenizer places more than 35\% of its word-internal boundaries on morpheme boundaries (boundary F1 at most 0.48). In a companion paper (arXiv:2611.13059) we present a tokenizer that reduces this tax. Code, pinned artifacts and raw token counts are available at https://github.com/HavonAI/indic-token-tax.
\end{abstract}


\section{Introduction}
Large language model (LLM) APIs charge per input and output token, and model context windows are measured in tokens. Tokenizers are usually trained on corpora dominated by English, so text in other languages tends to be split into more, smaller pieces. Speakers of those languages then pay more for the same request, wait longer for the same answer, and fit less of their text into the same context window.

This effect was documented in 2023 for commercial and open tokenizers across many languages~\cite{petrov2023unfairness,ahia2023allcost}. Since then, newer tokenizers with much larger vocabularies have been released, including OpenAI's o200k\_base (used from GPT-4o onwards) and models built specifically for Indian languages by Sarvam AI. Romanized Indian-language text, such as Hinglish (Hindi typed in Latin letters, often mixed with English), which is how a large share of Indian users actually type, is rarely measured.

This paper makes four contributions:
\begin{enumerate}
\item An updated, reproducible measurement of token disparity across nine public tokenizers, including four Sarvam tokenizers released in 2024 to 2026, for every language of India in FLORES-200 (25 language and script pairs).
\item Measurements on romanized text: a parallel Hinglish-English benchmark (PHINC), and romanized versus native-script token cost for 12 South Asian languages on about 10,000 human-romanized strings each (Dakshina).
\item A worked rupee cost example based on official per-token prices.
\item A morpheme boundary alignment measurement for Hindi and Marathi against an openly licensed segmentation set.
\end{enumerate}
A companion paper~\cite{arimbur2026tokenizer} presents a tokenizer trained to reduce the tax measured here; this paper evaluates only publicly released tokenizers.

\section{Related Work}
Tamang and Bora~\cite{tamang2024indian} compared 12 LLM tokenizers on the 22 official languages of India. Schmidt et al.~\cite{schmidt2024more} showed that fewer tokens do not by themselves make a better model, which is why we report compression only. MUTANT~\cite{rana2025mutant} builds a tokenizer for 22 Indian languages, English and code with language-aware pre-tokenization and multi-word aware training; its tokenizer was not publicly available for us to measure. Petrov et al.~\cite{petrov2023unfairness} showed that the same text can need up to 15 times more tokens across languages, and called for multilingually fair tokenizers. Ahia et al.~\cite{ahia2023allcost} showed that this translates into unequal API costs and lower utility for many languages. Rust et al.~\cite{rust2021tokenizer} found that tokenizer quality strongly affects multilingual model performance. Byte-pair encoding~\cite{sennrich2016bpe} and its byte-level variant~\cite{radford2019gpt2} are the basis of most current tokenizers, including all those measured here. For Indian languages, AI4Bharat's IndicLLMSuite~\cite{khan2024indicllmsuite} released the Sangraha corpus, and IndicTrans2~\cite{gala2023indictrans2} released the IN22 benchmarks. L3Cube released large romanized Hinglish~\cite{nayak2022hingcorpus} and Marathi-English~\cite{chavan2023mecorpus} corpora. MorphTok~\cite{brahma2025morphtok} proposed morphology-aware pre-tokenization for Hindi and Marathi and released the segmentation data we use in Section~\ref{sec:morph}.

\section{Methodology}
\subsection{Data}
\textbf{FLORES-200}~\cite{nllb2022,goyal2022flores}: 2,009 sentences (dev 997, devtest 1,012). FLORES-200 provides professionally translated, parallel sentences aligned with English, enabling cross-language token-count comparisons. Its documentation notes that several languages were translated through Spanish, French, Russian or Modern Standard Arabic rather than directly from English, without listing which; we did not verify the source language of each Indian-language subset. Our comparisons rely only on sentence alignment with English. We use every language of India present in the original release: Assamese, Awadhi, Bengali, Bhojpuri, Chhattisgarhi, Gujarati, Hindi, Kannada, Kashmiri (Arabic and Devanagari scripts), Magahi, Maithili, Malayalam, Marathi, Meitei (Bengali script), Mizo, Nepali, Odia, Punjabi, Sanskrit, Santali (Ol Chiki script), Sindhi (Arabic script), Tamil, Telugu and Urdu. Bodo, Dogri and Konkani, which are scheduled languages of India, are not in FLORES-200. Being parallel, the set allows comparing token counts for the same meaning. We used the original release, pinned by SHA-256 checksum.

\textbf{PHINC}~\cite{srivastava2020phinc}: 13,732 pairs of Hinglish social media posts and their human English translations after cleaning. It is parallel and in the same informal genre on both sides.

\textbf{L3Cube HingCorpus}~\cite{nayak2022hingcorpus} and \textbf{MeCorpus}~\cite{chavan2023mecorpus}: real romanized Hindi and Marathi social media posts. We draw 5,000 posts from each with a fixed seed (42). These are not parallel, so we report tokens per word only.

\textbf{MorphTok}~\cite{brahma2025morphtok}: word-level morpheme segmentations for Hindi (72,257 rows) and Marathi (61,736 rows), CC-BY-4.0, pinned to a Hugging Face revision. The authors built it by generating stem and affix combinations from affix lists compiled with language experts and by having experts post-edit the output of an existing segmenter~\cite{brahma2025morphtok}.

\textbf{Excluded: IN22-Conv}~\cite{gala2023indictrans2}, a conversational parallel benchmark for 22 Indian languages. Both its Hugging Face repository and the IN22 archive in the BPCC repository are gated and returned HTTP 401 without a logged-in account that has accepted the terms. We did not create an account, so conversational Devanagari text is not measured.

No text in this study was generated by a language model.

\subsection{Tokenizers}
We measure tiktoken cl100k\_base (GPT-4, GPT-3.5) and o200k\_base (GPT-4o and later OpenAI models); the Hugging Face tokenizers of Qwen2.5-7B-Instruct, DeepSeek-V3, Mistral-Nemo-Instruct-2407, Sarvam-1, Sarvam-M, Sarvam-30B and Sarvam-105B, all at pinned revisions. Claude, Gemini, Llama~3 and Gemma~3 are excluded: the first two have no public tokenizer, and the last two require gated access.

\subsection{Metrics}
Each text is encoded alone, with no special tokens or chat templates. For a language $L$, the \emph{parity ratio} is
\begin{equation}
P_L = \frac{\sum_i T(x_i^{L})}{\sum_i T(x_i^{\mathrm{en}})},
\end{equation}
where $T$ is the token count and $x_i^{L}$, $x_i^{\mathrm{en}}$ are aligned sentences. At equal per-token prices, $P_L$ is the cost multiplier for the same content, and a context window holds $1/P_L$ as much text. 95\% confidence intervals come from 1,000 bootstrap resamples of sentences.

Because $P_L$ divides by each tokenizer's \emph{own} English count, a tokenizer that is weak at English looks better than it is. We therefore also report a \emph{common-baseline} ratio, dividing by GPT-4o's English count on the same sentences.

\section{Results}
\subsection{Marathi, Hindi and Hinglish}
\begin{figure*}[t]
\centering
\includegraphics[width=0.92\textwidth]{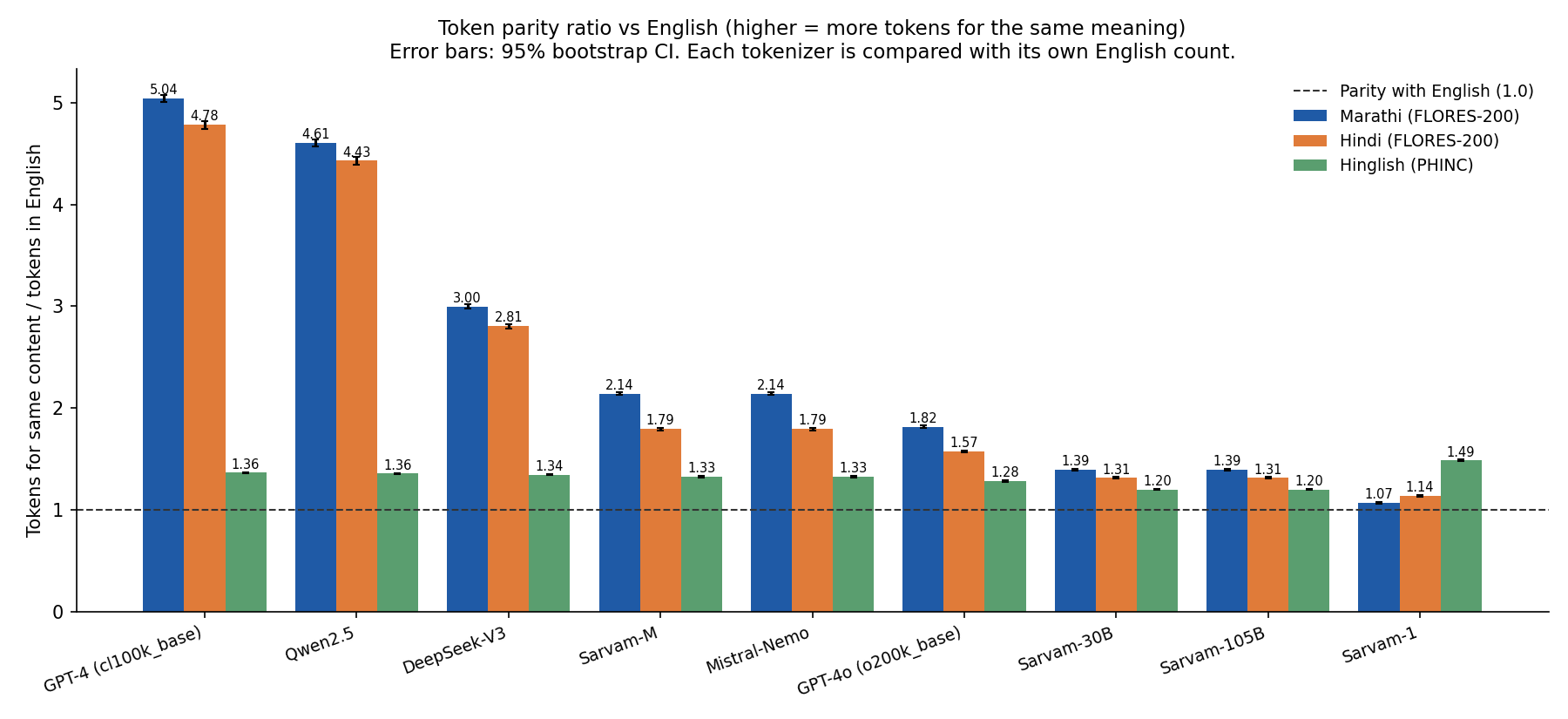}
\caption{Parity ratio versus English for each tokenizer, with 95\% bootstrap confidence intervals. The dashed line marks parity with English.}
\label{fig:parity}
\end{figure*}

Table~\ref{tab:parity} and Fig.~\ref{fig:parity} give the parity ratios. All 95\% intervals lie within $\pm0.041$ of the point estimates, and dev and devtest agree to within 0.035.

\begin{table}[t]
\centering
\caption{Parity ratio versus English (own English baseline). Lower is better.}
\label{tab:parity}
\begin{tabular}{lrccc}
\toprule
Tokenizer & Vocab & Marathi & Hindi & Hinglish \\
\midrule
GPT-4 (cl100k\_base) & 100,277 & 5.04 & 4.78 & 1.36 \\
Qwen2.5 & 151,665 & 4.61 & 4.43 & 1.36 \\
DeepSeek-V3 & 128,815 & 3.00 & 2.81 & 1.34 \\
Mistral-Nemo & 131,072 & 2.14 & 1.79 & 1.33 \\
Sarvam-M & 131,072 & 2.14 & 1.79 & 1.33 \\
GPT-4o (o200k\_base) & 200,019 & 1.82 & 1.57 & 1.28 \\
Sarvam-30B & 262,144 & 1.39 & 1.31 & 1.20 \\
Sarvam-105B & 262,144 & 1.39 & 1.31 & 1.20 \\
Sarvam-1 & 68,096 & 1.07 & 1.14 & 1.49 \\
\bottomrule
\end{tabular}
\end{table}

The tax is large for older and non-Indic-focused tokenizers: 4.6 to 5.0 times for Marathi on cl100k\_base and Qwen2.5. On these two tokenizers Devanagari averages about one character per token, meaning they mostly fall back to single characters or bytes. GPT-4o's o200k\_base reduces the Marathi tax from 5.04 to 1.82 times, and Hindi from 4.78 to 1.57 times. Short everyday sentences can look nearly free (``\raisebox{-0.25ex}{\includegraphics[height=1.1em]{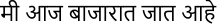}}'' is 6 tokens), but on full sentences Marathi still costs 82\% more than English. Sarvam-M produced token counts identical to Mistral-Nemo, its base tokenizer, and Sarvam-30B and 105B share one tokenizer file.

\subsection{Common English baseline}
Table~\ref{tab:common} corrects for differences in English compression. Sarvam-1 needs 17\% more tokens than GPT-4o for the same English text, which lowers its own-baseline parity.

\begin{table}[t]
\centering
\caption{Tokens per FLORES sentence divided by GPT-4o's English count (26.09). Lower is better.}
\label{tab:common}
\begin{tabular}{lccc}
\toprule
Tokenizer & English & Marathi & Hindi \\
\midrule
GPT-4o (o200k\_base) & 1.00 & 1.82 & 1.57 \\
GPT-4 (cl100k\_base) & 1.01 & 5.10 & 4.84 \\
Qwen2.5 & 1.03 & 4.74 & 4.55 \\
DeepSeek-V3 & 1.01 & 3.02 & 2.82 \\
Mistral-Nemo / Sarvam-M & 1.04 & 2.22 & 1.86 \\
Sarvam-30B / 105B & 1.01 & 1.41 & 1.33 \\
Sarvam-1 & 1.17 & 1.25 & 1.33 \\
\bottomrule
\end{tabular}
\end{table}

On this view, the best Indic-focused tokenizers (Sarvam-1, Sarvam-30B/105B) need 1.25 to 1.41 times the tokens GPT-4o needs for English to write Marathi and Hindi, so a gap remains even for tokenizers built for these languages.

\subsection{All languages of India in FLORES-200}
\begin{figure*}[t]
\centering
\includegraphics[width=\textwidth]{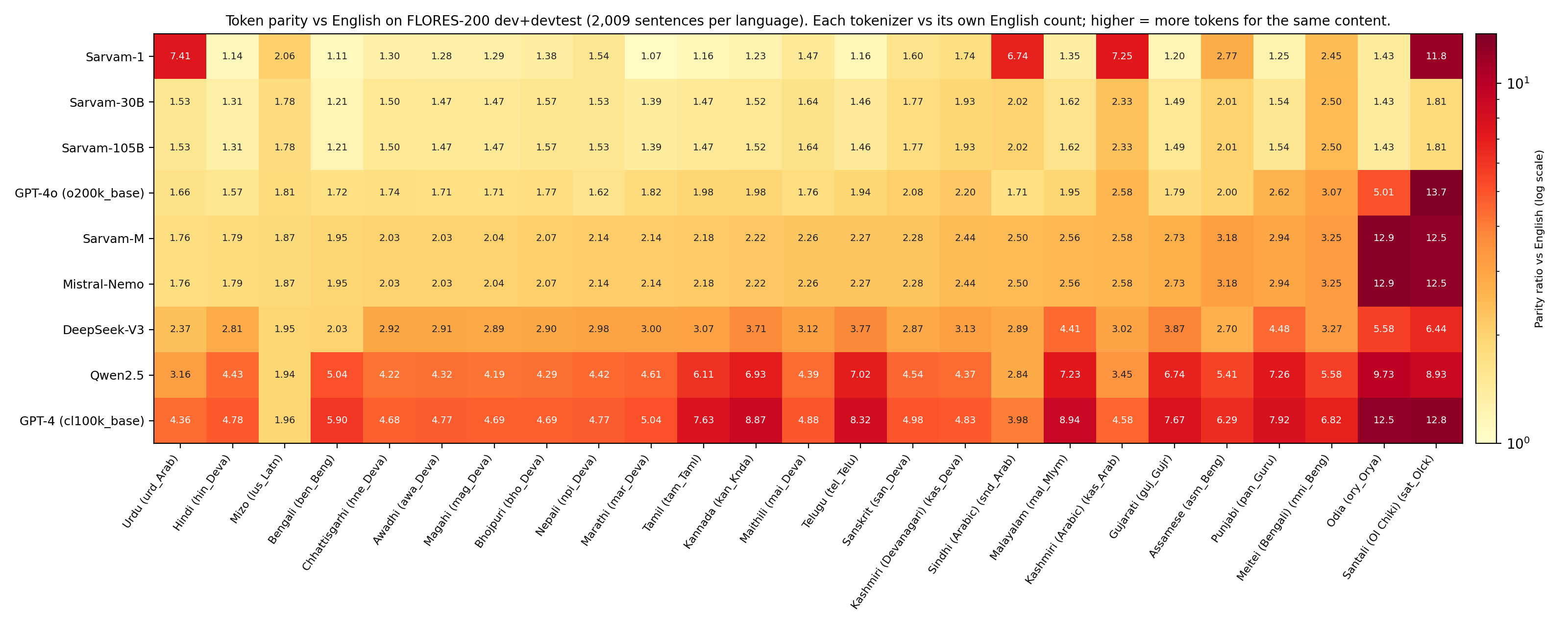}
\caption{Parity ratio versus English on FLORES-200 for every tokenizer (rows, sorted by median) and language of India (columns, sorted by median), on a log color scale.}
\label{fig:indic}
\end{figure*}

Table~\ref{tab:indic} and Fig.~\ref{fig:indic} extend the parity measurement to all 25 language and script pairs. The 95\% intervals lie within $\pm0.13$ of the point estimates for every cell.

\begin{table*}[t]
\centering
\caption{Parity ratio versus English (own English baseline) on FLORES-200 dev+devtest, 2,009 sentences per language, sorted by GPT-4o. Bold: lowest among the public tokenizers. Mistral-Nemo and Sarvam-M give identical counts, as do Sarvam-30B and Sarvam-105B.}
\label{tab:indic}
\footnotesize
\setlength{\tabcolsep}{3.5pt}
\begin{tabular}{llccccccc}
\toprule
Language & FLORES code & GPT-4 & Qwen2.5 & DeepSeek-V3 & Mistral-Nemo & GPT-4o & Sarvam-1 & Sarvam-30B \\
 & & cl100k & & & / Sarvam-M & o200k & & / 105B \\
\midrule
Hindi & hin\_Deva & 4.78 & 4.43 & 2.81 & 1.79 & 1.57 & \textbf{1.14} & 1.31 \\
Nepali & npi\_Deva & 4.77 & 4.42 & 2.98 & 2.14 & 1.62 & 1.54 & \textbf{1.53} \\
Urdu & urd\_Arab & 4.36 & 3.16 & 2.37 & 1.76 & 1.66 & 7.41 & \textbf{1.53} \\
Awadhi & awa\_Deva & 4.77 & 4.32 & 2.91 & 2.03 & 1.71 & \textbf{1.28} & 1.47 \\
Sindhi (Arabic) & snd\_Arab & 3.98 & 2.84 & 2.89 & 2.50 & \textbf{1.71} & 6.74 & 2.02 \\
Magahi & mag\_Deva & 4.69 & 4.19 & 2.89 & 2.04 & 1.71 & \textbf{1.29} & 1.47 \\
Bengali & ben\_Beng & 5.90 & 5.04 & 2.03 & 1.95 & 1.72 & \textbf{1.11} & 1.21 \\
Chhattisgarhi & hne\_Deva & 4.68 & 4.22 & 2.92 & 2.03 & 1.74 & \textbf{1.30} & 1.50 \\
Maithili & mai\_Deva & 4.88 & 4.39 & 3.12 & 2.26 & 1.76 & \textbf{1.47} & 1.64 \\
Bhojpuri & bho\_Deva & 4.69 & 4.29 & 2.90 & 2.07 & 1.77 & \textbf{1.38} & 1.57 \\
Gujarati & guj\_Gujr & 7.67 & 6.74 & 3.87 & 2.73 & 1.79 & \textbf{1.20} & 1.49 \\
Mizo & lus\_Latn & 1.96 & 1.94 & 1.95 & 1.87 & 1.81 & 2.06 & \textbf{1.78} \\
Marathi & mar\_Deva & 5.04 & 4.61 & 3.00 & 2.14 & 1.82 & \textbf{1.07} & 1.39 \\
Telugu & tel\_Telu & 8.32 & 7.02 & 3.77 & 2.27 & 1.94 & \textbf{1.16} & 1.46 \\
Malayalam & mal\_Mlym & 8.94 & 7.23 & 4.41 & 2.56 & 1.95 & \textbf{1.35} & 1.62 \\
Tamil & tam\_Taml & 7.63 & 6.11 & 3.07 & 2.18 & 1.98 & \textbf{1.16} & 1.47 \\
Kannada & kan\_Knda & 8.87 & 6.93 & 3.71 & 2.22 & 1.98 & \textbf{1.23} & 1.52 \\
Assamese & asm\_Beng & 6.29 & 5.41 & 2.70 & 3.18 & \textbf{2.00} & 2.77 & 2.01 \\
Sanskrit & san\_Deva & 4.98 & 4.54 & 2.87 & 2.28 & 2.08 & \textbf{1.60} & 1.77 \\
Kashmiri (Devanagari) & kas\_Deva & 4.83 & 4.37 & 3.13 & 2.44 & 2.20 & \textbf{1.74} & 1.93 \\
Kashmiri (Arabic) & kas\_Arab & 4.58 & 3.45 & 3.02 & 2.58 & 2.58 & 7.25 & \textbf{2.33} \\
Punjabi & pan\_Guru & 7.92 & 7.26 & 4.48 & 2.94 & 2.62 & \textbf{1.25} & 1.54 \\
Meitei (Bengali) & mni\_Beng & 6.82 & 5.58 & 3.27 & 3.25 & 3.07 & \textbf{2.45} & 2.50 \\
Odia & ory\_Orya & 12.48 & 9.73 & 5.58 & 12.93 & 5.01 & 1.43 & \textbf{1.43} \\
Santali (Ol Chiki) & sat\_Olck & 12.76 & 8.93 & 6.44 & 12.45 & 13.72 & 11.78 & \textbf{1.81} \\
\midrule
Median & & 4.98 & 4.54 & 3.00 & 2.26 & 1.82 & 1.38 & 1.53 \\
\bottomrule
\end{tabular}
\end{table*}

Four patterns stand out. First, on GPT-4o's tokenizer no language of India reaches parity: the lowest ratio is Hindi at 1.57, eight of the 25 are at 2.0 or above, and three scripts are much worse: Odia (5.01), Meitei in Bengali script (3.07) and Santali in Ol Chiki (13.72, higher than GPT-4's 12.76). Odia costs almost three times what Bengali costs (1.72) on the same tokenizer. Second, cl100k\_base and Qwen2.5 need 2.8 to 9 times for every language except Mizo, and 8.9 to 12.8 times on Odia and Santali, and DeepSeek-V3 needs 1.95 to 6.44 times; Mistral-Nemo stays between 1.76 and 3.25 except on Odia and Santali, where it exceeds 12. Third, the Sarvam tokenizers are the most even: Sarvam-30B/105B stays between 1.21 and 2.50 on every language, the only third-party tokenizer under 2.6 everywhere. Sarvam-1 has the lowest own-baseline ratio on 17 languages but needs 6.7 to 7.4 times for the three Arabic-script languages (Urdu, Sindhi, Kashmiri) and 11.8 times for Santali. Its own-baseline ratios are also flattered by its English, which needs 17\% more tokens than GPT-4o's; on the common baseline, Sarvam-1 leads Sarvam-30B by at most 0.16 on Marathi, Gujarati, Punjabi and the four Dravidian languages, ties it on Hindi (within 0.004), and trails it on the other 17, among them Bengali and Odia. Fourth, Mizo, written in Latin script, sits at 1.8 to 2.1 on every tokenizer, so Latin script alone does not remove the tax for a language with little web text.

\subsection{Romanized text}
Table~\ref{tab:roman} reports tokens per word on romanized text. PHINC is our primary evidence for romanized text because it is parallel; the HingCorpus and MeCorpus figures are secondary. On PHINC, Hinglish parity is low on every tokenizer (Table~\ref{tab:parity}), because Latin-script Hinglish reuses English sub-words. Part of the remaining gap is not a tokenizer effect: PHINC Hinglish posts contain 8\% more words than their English translations. Sarvam-1, the most efficient on Devanagari, is the least efficient on romanized text.

\begin{table}[t]
\centering
\caption{Tokens per word on romanized and social media text. Lower is better.}
\label{tab:roman}
\begin{tabular}{lcccc}
\toprule
 & \multicolumn{2}{c}{PHINC} & Hing- & Me- \\
Tokenizer & English & Hinglish & Corpus & Corpus \\
\midrule
GPT-4o & 1.36 & 1.61 & 1.45 & 2.14 \\
GPT-4 (cl100k) & 1.40 & 1.77 & 1.60 & 2.35 \\
Qwen2.5 & 1.41 & 1.77 & 1.59 & 2.37 \\
DeepSeek-V3 & 1.41 & 1.75 & 1.58 & 2.32 \\
Mistral-Nemo / Sarvam-M & 1.44 & 1.76 & 1.64 & 2.31 \\
Sarvam-30B / 105B & 1.38 & 1.53 & 1.37 & 2.11 \\
Sarvam-1 & 1.63 & 2.24 & 2.05 & 2.92 \\
\bottomrule
\end{tabular}
\end{table}

\subsection{Romanized versus native script (Dakshina)}
Dakshina~\cite{roark2020dakshina} provides, for 12 South Asian languages, 10,000 Wikipedia strings per language romanized by native-speaker annotators, so the same content is available in native script and in Latin letters. We report the ratio of tokens for the romanization to tokens for the native-script string (Table~\ref{tab:dakshina}); this compares two ways of writing the same text, not a language with English.

\begin{table*}[t]
\centering
\caption{Romanized versus native script on Dakshina: tokens for the human romanization divided by tokens for the same string in native script (below 1: romanized is cheaper). 10,000 strings per language (Urdu 9,759, Sindhi 9,999); all 95\% bootstrap intervals within $\pm0.01$.}
\label{tab:dakshina}
\footnotesize
\setlength{\tabcolsep}{3pt}
\begin{tabular}{lcccccccccccc}
\toprule
Tokenizer & Hin & Mar & Ben & Guj & Pan & Tam & Tel & Kan & Mal & Urd & Snd & Sin \\
\midrule
GPT-4o (o200k) & 1.08 & 0.97 & 0.96 & 1.03 & 0.70 & 1.04 & 0.90 & 0.95 & 1.00 & 1.10 & 1.07 & 0.70 \\
GPT-4 (cl100k) & 0.40 & 0.39 & 0.32 & 0.28 & 0.26 & 0.31 & 0.25 & 0.25 & 0.25 & 0.48 & 0.53 & 0.25 \\
Qwen2.5 & 0.43 & 0.43 & 0.37 & 0.32 & 0.28 & 0.39 & 0.29 & 0.32 & 0.31 & 0.67 & 0.75 & 0.31 \\
DeepSeek-V3 & 0.68 & 0.64 & 0.88 & 0.55 & 0.45 & 0.77 & 0.54 & 0.58 & 0.50 & 0.88 & 0.75 & 0.46 \\
Mistral-Nemo & 1.00 & 0.87 & 0.89 & 0.76 & 0.66 & 1.01 & 0.84 & 0.92 & 0.83 & 1.11 & 0.83 & 0.17 \\
Sarvam-30B/105B & 1.23 & 1.20 & 1.37 & 1.22 & 1.09 & 1.36 & 1.12 & 1.24 & 1.15 & 1.16 & 0.92 & 0.15 \\
Sarvam-1 & 1.76 & 1.75 & 1.64 & 1.70 & 1.58 & 1.86 & 1.63 & 1.76 & 1.55 & 0.30 & 0.33 & 0.20 \\
\bottomrule
\end{tabular}
\end{table*}

On the older or non-Indic-focused tokenizers, romanizing cuts token cost sharply: on cl100k\_base the romanized text needs only 0.25 to 0.53 of the native-script tokens, and on Qwen2.5 0.28 to 0.75, so typing in Latin letters avoids most of the tax these tokenizers impose on native scripts. On GPT-4o's o200k\_base the two are close (0.90 to 1.10 for most languages, but 0.70 for Punjabi and Sinhala). Indic-focused tokenizers reverse the effect: romanized text costs more than native script on Sarvam-30B/105B (1.09 to 1.37 for ten of the eleven Indian languages; Sindhi is 0.92) and Sarvam-1 (1.55 to 1.86 for the nine Brahmic-script languages of India). Dakshina's strings are carefully romanized formal text, so casual messaging may tokenize differently.

\subsection{What the tax costs in rupees}
\label{sec:cost}
To turn parity into money we take official list prices for models whose tokenizer we measured (Table~\ref{tab:prices}), accessed on 5 November 2026, standard tier, with no caching, batch or off-peak discounts. OpenAI prices are from its API pricing page~\cite{openai2026pricing}; tiktoken 0.12.0, OpenAI's tokenizer library, maps gpt-4o, gpt-4.1-mini, gpt-5 and gpt-5-mini to o200k\_base. DeepSeek's API now serves DeepSeek-V4.1-Flash and DeepSeek-V4-Pro~\cite{deepseek2026pricing}; we compared the published \texttt{tokenizer.json} of DeepSeek-V4.1-Flash, V4-Flash and V4-Pro with DeepSeek-V3 component by component: vocabulary, merges, normalizer, pre-tokenizer, decoder and post-processor are identical (by SHA-256 of their canonical JSON), only the added special tokens differ, and all four encode a 1,400-sentence FLORES sample in seven languages to identical token ids, so the V3 counts apply to plain text. The comparison script and hashes are in the repository. Sarvam bills Sarvam 105B in rupees~\cite{sarvam2026pricing}; we assume the API uses the tokenizer of the open Sarvam-105B release, which we did not verify. USD prices are converted at INR 94.3996 per USD, the FBIL reference rate at 1:00 pm on 5 November 2026 as published by the Reserve Bank of India~\cite{rbi2026fx}. We did not use Qwen prices: the Qwen API models listed by Alibaba Cloud are not the Qwen2.5 release we measured, and we found no published statement that they share its tokenizer.

\begin{table}[t]
\centering
\caption{List prices per million tokens used in Section~\ref{sec:cost}, accessed 5 November 2026.}
\label{tab:prices}
\footnotesize
\setlength{\tabcolsep}{2.4pt}
\begin{tabular}{llrr}
\toprule
Model (API) & Tokenizer & Input & Output \\
\midrule
GPT-4o & o200k\_base & USD 2.50 & USD 10.00 \\
GPT-4.1 mini & o200k\_base & USD 0.40 & USD 1.60 \\
GPT-5 & o200k\_base & USD 1.25 & USD 10.00 \\
GPT-5 mini & o200k\_base & USD 0.25 & USD 2.00 \\
deepseek-flash$^a$ & DeepSeek-V3 & USD 0.30 & USD 1.20 \\
deepseek-v4-pro$^a$ & DeepSeek-V3 & USD 1.32 & USD 3.96 \\
Sarvam 105B & Sarvam-105B & INR 15.00 & INR 60.00 \\
\bottomrule
\multicolumn{4}{l}{$^a$Peak-hour cache-miss price; off-peak is half.}
\end{tabular}
\end{table}

The figures in this section are illustrative estimates, not measured production costs or savings. The worked example is a hypothetical customer support bot with 100,000 chats a month, each with 800 input and 400 output tokens of conversation text when held in English on that model's tokenizer. These two numbers are assumptions, not measurements. The same conversation in language $L$ is taken to cost the English cost times $P_L$ for both input and output, which assumes the reply is in the user's language and that FLORES parity carries over to support conversations. A fixed English system prompt would not be multiplied and would shrink the gap.

\begin{table}[t]
\centering
\caption{Worked example: monthly cost in thousands of INR for 100,000 chats (800 input and 400 output English-equivalent tokens per chat). Full results for all models and languages are in the released \texttt{cost.csv}.}
\label{tab:cost}
\footnotesize
\setlength{\tabcolsep}{2.8pt}
\begin{tabular}{lrrrr}
\toprule
Language & GPT-4o & GPT-5 mini & deepseek-flash & Sarvam 105B \\
\midrule
English & 56.6 & 9.4 & 6.8 & 3.6 \\
Hinglish (PHINC) & 72.6 & 12.1 & 9.1 & 4.3 \\
Hindi & 89.1 & 14.9 & 19.1 & 4.7 \\
Marathi & 102.8 & 17.1 & 20.4 & 5.0 \\
Bengali & 97.3 & 16.2 & 13.8 & 4.4 \\
Tamil & 112.0 & 18.7 & 20.9 & 5.3 \\
Telugu & 109.8 & 18.3 & 25.6 & 5.2 \\
Gujarati & 101.4 & 16.9 & 26.3 & 5.4 \\
Punjabi & 148.4 & 24.7 & 30.5 & 5.6 \\
Urdu & 94.0 & 15.7 & 16.1 & 5.5 \\
Odia & 283.6 & 47.3 & 37.9 & 5.1 \\
\bottomrule
\end{tabular}
\end{table}

Table~\ref{tab:cost} shows the result. On GPT-4o, the bot costs INR 56,640 a month in English, INR 89,145 in Hindi, INR 102,829 in Marathi and INR 283,601 in Odia, so the same service in Marathi costs INR 46,190 a month more than in English. On deepseek-flash, Marathi costs three times English (INR 20,384 versus 6,797). Differences in per-token price between models are larger than the tax itself (GPT-4o's English cost is 16 times Sarvam 105B's), so the tax matters most when comparing languages on the same model. These figures say nothing about answer quality, which we did not measure, and exclude taxes and chat-template overhead.

\subsection{Morpheme boundary alignment}
\label{sec:morph}
Fewer tokens is not the only property one might want: tokens that follow morpheme boundaries may also help models generalize~\cite{brahma2025morphtok}. We measure this on the MorphTok segmentations. We keep only entries whose segments concatenate exactly to the word, so that each gold boundary is a character offset (this drops sandhi entries with character changes: 10,491 Hindi and 19,348 Marathi rows), and drop words listed with conflicting segmentations. This leaves 59,676 Hindi words (46,144 gold boundaries) and 42,292 Marathi words (39,083 boundaries). Each word is encoded alone after a space, as it would appear mid-sentence. A predicted boundary is a character offset inside the word where one token ends and the next begins; splits inside a single character, from byte fallback, do not count. We report micro-averaged boundary precision, recall and F1. As a reference, splitting after every character gives recall 1 and a precision equal to the density of gold boundaries, which is what a tokenizer with no knowledge of morphology would reach at that fragmentation.

\begin{table}[t]
\centering
\caption{Morpheme boundary precision (P), recall (R) and F1 on MorphTok, and tokens per word (T/w). Higher P and F1 are better.}
\label{tab:morph}
\footnotesize
\setlength{\tabcolsep}{2.7pt}
\begin{tabular}{lcccc|cccc}
\toprule
 & \multicolumn{4}{c|}{Hindi} & \multicolumn{4}{c}{Marathi} \\
Tokenizer & P & R & F1 & T/w & P & R & F1 & T/w \\
\midrule
GPT-4 (cl100k)$^a$ & .13 & .72 & .22 & 7.96 & .19 & .98 & .32 & 9.08 \\
DeepSeek-V3 & .14 & .70 & .24 & 4.99 & .20 & .95 & .33 & 5.65 \\
Mistral-Nemo$^b$ & .18 & .65 & .29 & 3.83 & .24 & .88 & .37 & 4.51 \\
GPT-4o (o200k) & .20 & .62 & .31 & 3.37 & .28 & .85 & .43 & 3.79 \\
Sarvam-1 & .29 & .79 & .43 & 3.25 & .34 & .73 & .46 & 3.01 \\
Sarvam-30B/105B & .34 & .79 & .48 & 2.78 & .34 & .79 & .48 & 3.15 \\
\midrule
Every character & .14 & 1.00 & .25 & 6.40 & .15 & 1.00 & .26 & 7.26 \\
\bottomrule
\multicolumn{9}{l}{$^a$Qwen2.5: same P and R; T/w 7.16 (Hindi), 8.16 (Marathi).}\\
\multicolumn{9}{l}{$^b$Sarvam-M gives identical results.}
\end{tabular}
\end{table}

Table~\ref{tab:morph} shows that alignment is weak everywhere. cl100k\_base, Qwen2.5 and DeepSeek-V3 have precision at or near the every-character reference (0.13 to 0.20), so their high recall comes from fragmentation rather than morphology. The Indic-focused Sarvam tokenizers place 29 to 34\% of their word-internal boundaries on morpheme boundaries, about twice the reference, and no tokenizer exceeds an F1 of 0.48. Sarvam-30B/105B aligns best on both languages (F1 0.48; exact segmentation match on 27\% of Hindi and 15\% of Marathi words). Low fragmentation and alignment go together here, but neither implies the other.

\section{Discussion}
For a team choosing an API for workloads in Indian languages, the tokenizer alone can change token cost by a factor of two to more than ten for the same content, and the factor depends on the script as much as on the model: GPT-4o is reasonable for Hindi and Bengali but expensive for Odia and Santali. This should be checked per language before comparing per-token prices. The newest OpenAI tokenizer narrows but does not close the gap. Indic-focused tokenizers come close to parity on the scripts they were built for, and Sarvam-30B/105B is the most even across all 25. Sarvam-1 shows that a 68,096-token vocabulary can be competitive on the scripts it was trained for, but it needs 6.7 to 11.8 times the English token count on the three Arabic-script languages and Santali. Token count also says nothing about model quality.

\section{Limitations}
(1) We did not check FLORES sentences for overlap with the web-crawled training data of the tokenizers measured. (2) Romanized text has no English-aligned parallel set beyond PHINC (Hinglish): Dakshina pairs romanizations with native script, not with English, and its formal Wikipedia strings may not reflect casual typing, so MeCorpus tokens per word remain our only measurement of informal romanized Marathi. (3) FLORES is formal, Wikipedia-style text, and conversational text may behave differently; IN22-Conv, which would test this, is gated and was not used. (4) Romanized and morphological measurements cover only Hindi and Marathi, and Bodo, Dogri and Konkani are absent from FLORES-200. (5) We count raw text tokens only, excluding chat-template overhead. (6) L3Cube HingCorpus and MeCorpus are licensed CC BY-NC-SA 4.0 for research purposes; our measurements are research use. HingCorpus was read from a third-party Hugging Face mirror (hinglish-orthobench/HingCorpus) of the L3Cube release. (7) We use the original FLORES-200 release, not the corrected FLORES+ continuation, and did not audit translation quality for each language. (8) The rupee example rests on assumed chat lengths, list prices that change often, the assumption that output parity equals input parity, and, for Sarvam, an unverified assumption about which tokenizer its API uses; the OpenAI model to tokenizer mapping comes from tiktoken rather than an API document. (9) MorphTok segmentations are partly generated from stem and affix lists and partly post-edited model output, so they are a curated reference rather than an exhaustive gold standard; we score single words out of context and drop entries with sandhi changes.

\section{Conclusion}
Indian-language users still pay a token tax in 2026. On GPT-4o's tokenizer it is 1.6 to 2.6 times for 22 of the 25 languages of India, 3.1 times for Meitei and 5 to 14 times for Odia and Santali; on several widely used open tokenizers it is 3 to 9 times. In an illustrative support-bot scenario of 100,000 chats a month on GPT-4o, Marathi would cost roughly INR 46,000 a month more than English. Indic-focused tokenizers reduce the tax substantially, Sarvam-30B/105B most evenly across scripts, but token boundaries still align poorly with morpheme boundaries on every tokenizer. Future work: conversational text through the gated IN22 benchmarks, and Claude via its token-counting API.

\section*{Reproducibility}
Code and data: \url{https://github.com/HavonAI/indic-token-tax}. The repository contains the measurement scripts (\texttt{measure.py}, \texttt{morph\_align.py}, \texttt{cost.py}, \texttt{dakshina.py}); the tokenizer equivalence check (\texttt{check\_tokenizer\_equivalence.py}) with its SHA-256 output; the pinned FLORES-200 checksum and Hugging Face revision identifiers of every dataset and tokenizer; the 5,000-post sampling seed; the bootstrap procedure and confidence-interval outputs; raw per-language token counts (\texttt{results/results.csv}, \texttt{indic\_parity.csv}); the price sources.

\end{document}